\RequirePackage{fix-cm}
\documentclass[smallcondensed]{svjour3}     
\smartqed  
\usepackage{graphicx}
\usepackage{amssymb}
\usepackage{amsmath}
\usepackage{amsfonts}
\usepackage{hyperref}
\usepackage{chapterbib}
\usepackage{geometry}
\usepackage{multirow}

\journalname{Journal of Intelligent Information Systems}

\begin{document}

\title{Case Studies on using Natural Language Processing Techniques in Customer Relationship Management Software
}

\titlerunning{Case Studies on using NLP in CRM software}        

\author{Şükrü Ozan \href{https://orcid.org/0000-0002-3227-348X}{\includegraphics[scale=0.075]{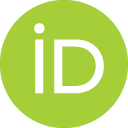}} }

\authorrunning{Ş. Ozan} 

\institute{Şükrü Ozan \at
              AdresGezgini Inc. Research \& Development Center Bayraklı, İzmir, TURKEY \\
              Tel.: +90-232-441-40-11\\
              Fax: +900-232-441-40-12\\
              \email{sukruozan@adresgezgini.com}           
}

\date{Received: date / Accepted: date}

\maketitle

\begin{abstract}
How can a text corpus stored in a customer relationship management (CRM) database be used for data mining and segmentation? In order to answer this question we inherited the state of the art methods commonly used in natural language processing (NLP) literature, such as word embeddings, and deep learning literature, such as recurrent neural networks (RNN). We used the text notes from a CRM system which are taken by customer representatives of an internet ads consultancy agency between years 2009 and 2020.  We trained word embeddings by using the corresponding text corpus and showed that these word embeddings can not only be used directly for data mining but also be used in  RNN architectures, which are deep learning frameworks built with long short term memory (LSTM) units, for more comprehensive segmentation objectives. The results prove that structured text data in a CRM can be used to mine out very valuable information and any CRM can be equipped with useful NLP features once the problem definitions are properly built and the solution methods are conveniently implemented. 
\end{abstract}

\section{Introduction}\label{sec:intro}

CRM is a sine qua non of sales and marketing operations of most businesses. A good CRM software not only manages  relationship with the customers but also manages every related data  about them. Since recent advances in technology make it possible to store huge amount of data, data mining in CRM has gained a significant popularity in related fields. 

In this study we concentrate on the structured text data populated in the CRM of an internet advertising consultancy agency called AdresGezgini Inc. (hereinafter referred to as ``the company") which is located in Turkey. The company extensively uses telephony since it is important not  only for marketing processes but also for customer relationship management. Telephone calls performed over call centers guarantee customer satisfaction and loyalty, if the services or products already meet the customer needs \cite{Feinberg:2000}.  

Especially for companies that prefer telephony as the major sales and marketing strategy, it is expected from the customer representatives (a.k.a. the agents) that they take notes regarding customer calls. These notes reflects the idea of the agents about the corresponding call which also reflects the customers general attitude and ideas about the company and its services. Agents try to briefly explain the conversation they performed in their own words. It's  worthy to emphasize that the sensitive information of customers, such as contact information, monetary information etc., is not included  in the processed corpus. 

Since a customer representative makes tens of these calls a day,  it is not uncommon to have a large corpus of text data in CRM databases and we claim that text data stored in a CRM can be very useful. While making this claim we present three problems and show our solution approaches which are mainly based on recent techniques in NLP and machine learning (ML) literature. We especially hope to draw attention of NLP researchers and enthusiasts and lead their way in solving similar open ended problems in CRM literature and present three case studies where we introduce three problems and solve them by using state of the art NLP methods.

The article is organised as follows. The next section covers the related work in the literature. In Section \ref{sec:data} we  give as much detail as possible for the data used in the corresponding experiments, since we believe that understanding the data is the most important aspect to completely comprehend an artificial intelligence (AI) problem and its solution. Section \ref{sec:wordembed} is about the specifics of word embeddings, on which we base our learning algorithms. We start describing the problems and our solution approaches to these problems in Section \ref{sec:casestud}. In Section \ref{sec:implement} we give more implementation  details for enthusiastic readers who may want to try applying the proposed methods to their  data. The experimental results and concluding remarks are given in Section \ref{sec:results}.

\section{Related Work}\label{sec:related}

In this study we extensively refer to the recent studies in  NLP literature which can be used to seek solutions to some text related problems fictionalized to empower a CRM software. 

The most convenient way to make an AI algorithm conceive the meaning of a word is to use the state of the art method called word embeddings. The aim of this method is to meet the meaning of a word by using multi dimensional vectors. There are two major methods to find word embedding vectors namely Word2Vec \cite{Mikolov:2013} and GloVe \cite{Pennington:2014}. Majority of NLP problems are now solved by using  RNNs since the word streams (i.e. sentences) can be considered as  sequences where the information may travel along the temporal dimension in both forward and backward directions \cite{Schuster:1997}. Word embeddings are  used together with RNNs to solve complicated NLP problems like end-to-end machine translation. Such an algorithm  has been used in Google Translate since 2016 \cite{Google:2016} . 

Using word embeddings for text mining is a very effective method and it is used for text mining in various literature. A recent work \cite{Jiang:2015} addresses the application of this idea where the aim is to mine important medical information out from biomedical data . In \cite{Bahari:2015},  data mining is used to mine out useful data from CRM in order to model customer behaviour. Modeling customer behaviour makes it possible to predict future and reduce customer churn. In the first case study of this work, we inherit the same idea and apply it for text mining in text corpus of the agents' after call notes that are recorded in the CRM software.


 On the other hand, vast amount of recent studies are concentrated on using data mining to segment customers. The importance of customer segmentation in CRM  is addressed in  \cite{Thakur:2016}. It is possible to segment customers using cluster analysis that may be helpful to the marketers in increasing their profit, sales and extending  customer continuity \cite{Tsiptsis:2010}. In \cite{Gupta:2016} and \cite{Windler:2017}, the segmentation problem is investigated in a managemental point of view. 

However, it is also possible to segment customers in a monetary based aspect. In \cite{Sarvari:2016} recency, frequency and monetary (RFM) considerations together with demographic factors are used to perform customer segmentation. Similarly in a more recent work, an empirical study is performed to find out new strategies for customer segmentation in digital, social media and mobile marketing (DSMM) field where monetary data, such as buying frequency, is used \cite{Muller:2018}. 

Moreover, thanks to recent advances in AI, now it is possible to make a good use out of ML for such problems. In \cite{Ozan:2018} monetary based customer segmentation problem is addressed where payment information of customers are used for segmentation by using  traditional ML methods like linear and logistic regression. Also in \cite{Ozan:2019}  the same problem is solved by using a simple neural network and the accuracy is further improved. 


Segmentation related works given above extensively relies on direct customer data including personal and monetary details. In the second case study of this work, we rather use the notes taken by agents and segment the leads according to these notes.  `Lead' can be defined as the combination of both active and potential customers. This method requires the interpretation of the notes. For an AI algorithm to understand the meaning of a complete sentence it should be able to process temporal data. LSTM-RNNs have proven their superiority over standard RNNs to learn long-term dependencies \cite{Hochreiter:1997}. Since LSTM architectures are able to handle temporal data they are also commonly used in audio processing applications like acoustic feature extraction \cite{Lenglaive:2015}. In this study LSTM-RNN is used in NLP terms and for the proposed binary lead segmentation operation its proven to achieve good accuracy results. In the literature we also see that LSTM-RNN architecture is also used for sentiment classification from short sentences \cite{Novak:2017}, \cite{Wang:2018}. This idea inspired us to identify a person from written notes. Hence we slightly modified our training data used for lead segmentation and used it to identify agents from their notes. We also further trained this network structure as a word suggestion tool with a specific agent's text entries. We obtained promising results from these tasks and presented them in the proceeding chapters.


\section{Data}\label{sec:data}

The data consists of Turkish text notes taken by the customer representatives (a.k.a. agents) after the marketing or service calls they perform with their customers. The company uses its own CRM software running on a linux server where the data is stored in a MySQL v.5.6 database. There are more than one million different notes taken by agents from the year 2009 up to 2020. To be more specific; there are 314 unique agents and 1.066.086 notes.

It is a convenient approach to preprocess the data and tailor it such that it can be used in more enhanced processing steps like word embedding calculations. Common preprocessing operations can be listed as:

\begin{itemize}
    \itemsep0em     
    \item Making the sentences lower case,
    \item Making the sentences punctuation free,
    \item Removing repeated sentences,
    \item Leaving out outliers, i.e. do not consider extremely short and extremely long text entries.
\end{itemize}

\begin{figure}[h!]
  \centering
  \includegraphics[width=0.75\textwidth]{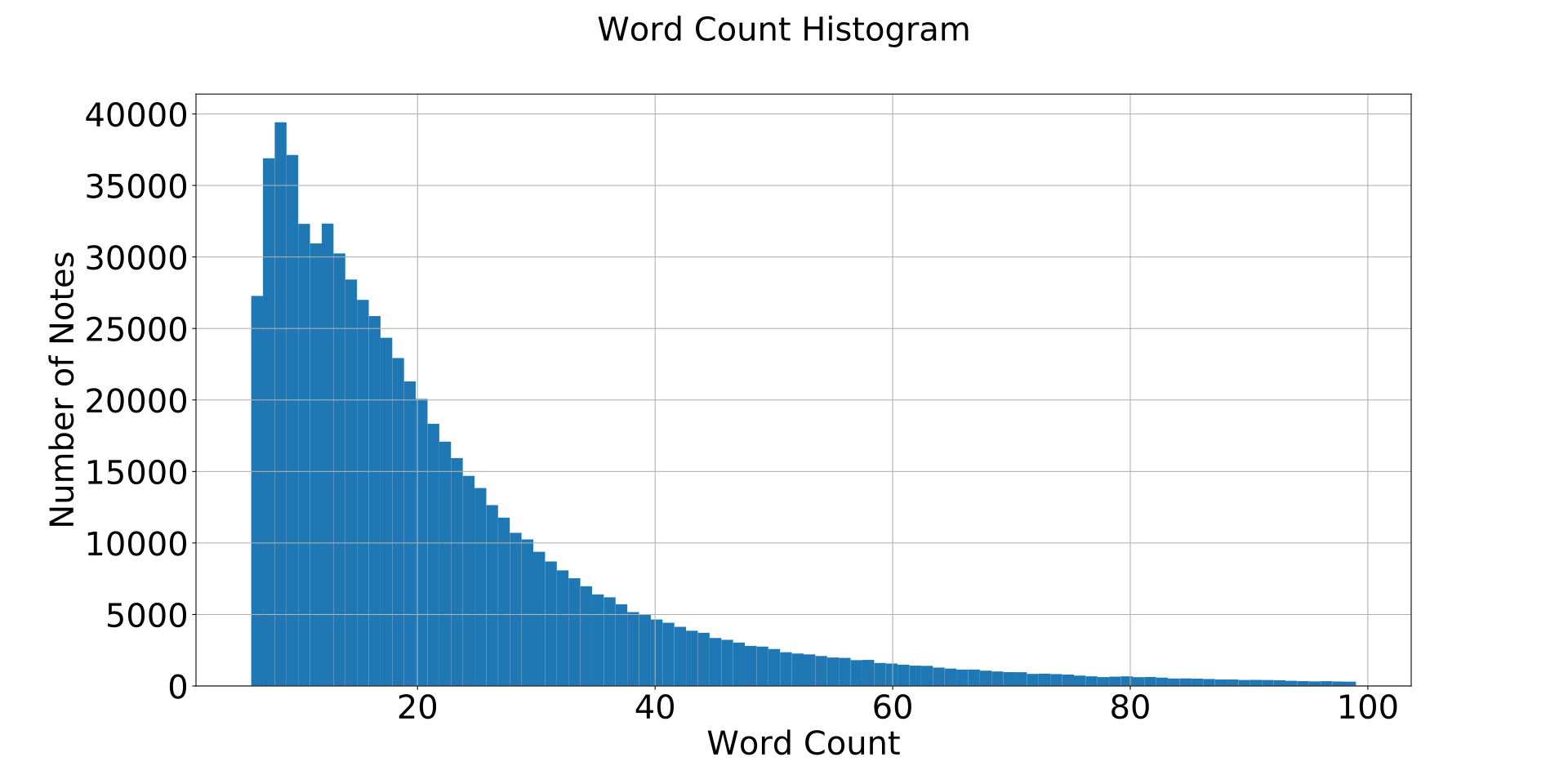}
  \caption{Histogram of the word counts of the notes in the database. For display purposes outliers are omitted and only the notes with word count greater than 5 and less than 100 are considered.}
  \label{fig:wc_histogram}
\end{figure}

After applying these preprocessing operations and determining individual word counts for each sentence, we obtain the distribution in Figure \ref{fig:wc_histogram}.  While plotting we omitted the outliers. The outliers are  extremely long  (longer than 100 words) and extremely short (shorter than 5 words  ) notes. There are not much long notes, but on the contrary there are quite a lot of short notes. These are very short notices like ``ulaşamadım" (I could not reach) and ``müsait değil" (customer is not available). 

 The statistical analysis of the remaining data can be seen in Table \ref{tab:datastats}. These statistics show  that the number of  sentences, (i.e. count) after removing the outliers and duplicate notes, is  approximately 800K and totals up to 17.599.462 words; the average number of word count (i.e. mean) is 21.4; standard deviation (i.e. std) is 16.2; 25\%, 50\% and 75\% reflects  the respective percentile values of the distribution. 

\begin{table}[h!]
    \centering
    \begin{tabular}{|c|c|c|c|c|c|c|}
        \hline
        \textbf{Stat}   &   count&    mean&   std&    25\%&   50\%&   75\%\\    \hline  
        \textbf{Value}  &   799868&   21.4&   16.2&   10.0&   17.0&   27.0\\      \hline
    \end{tabular}
    \caption{Statistical details of the word count data distribution shown in Figure \ref{fig:wc_histogram}. }
    \label{tab:datastats}
\end{table}

\begin{figure}[h!]
  \centering
  \includegraphics[width=0.60\textwidth]{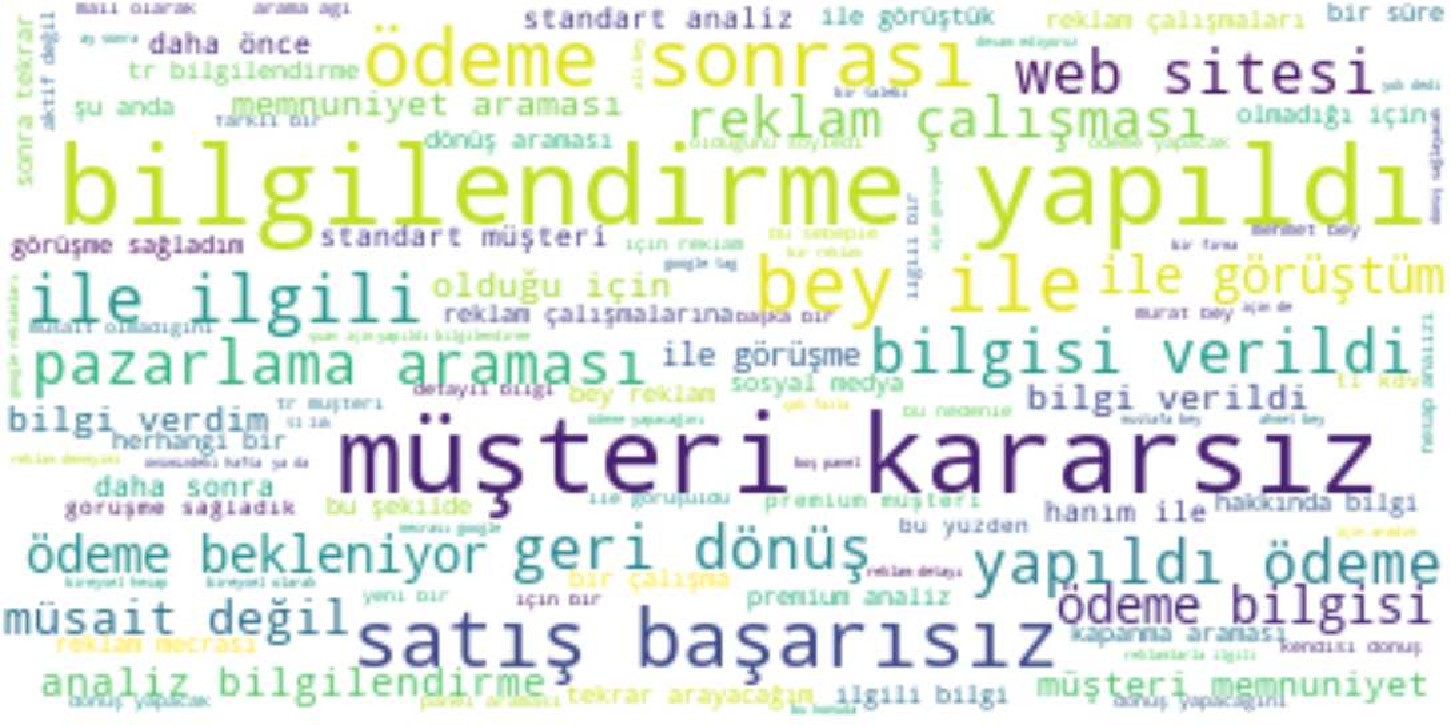}
  \caption{By using python word cloud library \cite{Mueller:2015} it is possible to create a cloud of words. In this figure we can also capture the words  frequently used together. E.g. ``bilgilendirme yapıldı" (notified), ``müşteri karasız" (the customer is indecisive), ``satış başarısız" (the sale has failed). }
  \label{fig:wordcloud}
\end{figure}

 In Figure \ref{fig:wordcloud} we show frequently used tuples where more frequent tuples have larger font and less frequent ones have smaller fonts. By looking at this word cloud, one can easily capture the phrases like ``bilgilendirme yapıldı" (notified), ``müşteri karasız" (the customer is indecisive), ``satış başarısız" (the sale has failed) etc., which are commonly used by agents during note taking.

It is also possible to process the text corpus of approximately 18M words and find out the frequency of each word. By using Gensim Library \cite{Rehurek:2010} we find the frequency of each word. This library is actually used to generate word embeddings, which we will present explicitly in the next section, by training on a given text corpus. It is a common strategy to omit words which are rarely used. We kept the minimum usage limit as five, i.e. we did not train on words which are used less than 5 times. 

In Table \ref{tab:wordfrequencies} we listed five most and least frequent words where we see the prefix ``bey" (mr.) as the most frequently used word. As least frequent words we see some brand names and the tag words ``UNK" and ``PAD". We will explain details of  how and why we use these tag words in Seciton \ref{sec:implement}.

\begin{table}[h!]
\centering
\begin{tabular}{ |c|c|c|c|c|c| } 
\hline
Group & \multicolumn{5}{|c|}{Word - Word Count } \\
\hline
\multirow{3}{4em}{Most Frequent Words } 
 & bey & için & bir & müşteri & ile \\ 
 & (mr.)    & (for)   & (a)    & (customer)    & (with) \\
 & 253296    & 236406    &  218995  & 216869    & 214831 \\
\hline
\multirow{3}{4em}{Least Frequent Words } 
 & luffi & wstark & hysafe & UNK & PAD \\ 
 & 5    & 5    & 5    & 1    & 1 \\
 &     &     &    &     &  \\
\hline
\end{tabular}
\caption{The table shows  five most frequently and five least frequently used words and their counts in the text corpus. Total number of words in the corpus is approximately 18M (i.e. 17.599.462) }
\label{tab:wordfrequencies}
\end{table}

Since  the leads are manually labelled with two major tags, i.e. high priority and low priority, each note regarding a lead can be labelled  with the same tag. Hence, it is possible to extrapolate unique labels for each sentence. Moreover, the corresponding agent id for any taken note is also known. Hence each note in the database can also be labelled with the corresponding agent id. With these two labelling schemes we later present  two case studies in Section \ref{sec:casestud}.
 

\section{Word Embeddings}\label{sec:wordembed}

NLP aims to find solutions to problems in natural language understanding such as syntax, semantics and discourse \cite{Bates:1995}. With the recent advances in the computer technology and ML theory very complicated NLP problems, like one to one translation of a text to any other language, can now be  solved with considerably high accuracies.

Representing words as high dimensional vectors is  the best way to capture their meanings. Converting a word to a numerical vector requires a probabilistic calculation considering the probability of each word representing its frequency in the whole language and  probabilities of these words being used together with other context words in a predefined neighbourhood. Word2Vec \cite{Mikolov:2013} and GloVe \cite{Pennington:2014} are the two main approaches to find vector representations of words. In this study Word2Vec approach is used. 

Size of the word vectors varies between 50 to 500 in practical applications. In most research studies like \cite{Bolukbasi:2016}, 300 is the most common choice for embedding size. On the contrary, in \cite{Jurafsky:2019}, it is recommended to use denser vectors rather than sparse vectors  . Depending on the size and complexity of training data choosing a smaller embedding size may yield better results since sparse vectors tend to overfit in deep learning applications.

\subsection{Word2Vec}\label{sec:word2vec}

In this section we consider Word2Vec in more detail in accordance with \cite{Jurafsky:2019}. This approach mainly relies on skip-grams. Skip-gram with negative sampling intuition can be given as :

\begin{enumerate}
    \item   Consider the word of interest and mark its neighbouring words within a predefined neighborhood as positive examples.
    \item   Select random words from the dictionary excluding positive examples and consider them as negative samples.
    \item   By using simple logistic regression, train a classifier to distinguish between positive and negative samples. 
    \item   Use the weights of the regression as the embeddings. 
\end{enumerate}

Let's explain the derivation of embedding scheme with an example sentence:

\begin{center}
...that dense vectors work better $\underbrace{in}_{c_1}$ $\underbrace{every}_{c_2}$ $\underbrace{NLP}_{t}$ $\underbrace{task}_{c_3}$ $\underbrace{than}_{c_4}$  sparse vectors...
\end{center}

In this example $t$ represents target word where  $c_i$s represent its context words in a predefined neighborhood where in this case its 2 words in both directions (i.e. there are 4 context words for any target word). The probability that a given word $c_i$ is a positive context word of $t$ can be represented as Equation \ref{eq:context}, hence for the same word the probability of being a negative context word of $t$ can be represented as Equation \ref{eq:noncontext}. The proposed algorithm for Word2Vec calculates these probabilities by using the similarity between two vectors representing the words $t$ and $c$ respectively. 

 \begin{align} 
 P(+|t,c) \label{eq:context}\\
 P(-|t,c) & = 1 - P(+|t,c) \label{eq:noncontext}
 \end{align}
 
 The probability in Equation \ref{eq:context} can be interpreted as the similarity between  $t$ and $c$. Similarity between two vectors  can be calculated by using their dot product. Since the dot product gives value between $-\infty$ and $\infty$ it is important to map it between 0 and 1 such that $P(.)$ becomes proper probability function. Sigmoid function is the best choice for the corresponding mapping operation. 

 
 
 
 

Word2Vec method first assigns randomly initialized embedding vectors to each word in the dictionary. An iterative process is performed to align these embedding vectors to their optimal positions such that these vectors represent their frequency of being used together with other context words as much as possible. 

If we consider the sample sentence above, there are 4 positive context words or a target word. If we choose ``NLP" as the target word ``in", ``every", ``task" and ``that" become positive context words in the predefined neighborhood window ( say $W=\pm 2$ words) of the target word. The aim is to train a binary classifier which classifies a given context word into either positive or negative class. Positive context words are the words in the predefined neighborhood, but for a proper training we also need negative samples and they can be populated by using randomly selected words from the lexicon except the ones used within the neighborhood of the target word. The corresponding classifier uses a learning objective  $L(\theta)$ which can be written as Equation \ref{eq:learnobj2}, where $k$ is the number of randomly selected negative context words for a target word $t$.

\begin{align} 
L(\theta) &=  log\,P(+|t,c) + \sum_{i=1}^k log\,P(-|t,n_i)\nonumber\\
          &=  log\,\frac{1}{1+e^{-c\cdot t}}+\sum_{i=1}^k log\,\frac{1}{1+e^{n_i\cdot t}}\label{eq:learnobj2}
\end{align}

Word2Vec scheme aims to maximize the dot product of the target word vector  with its positive context words and minimize the dot product of the target word vector with its negative context words. It is an iterative process and the number of iterations can be determined in Gensim software library .

\subsection{Sanity Check}\label{sec:sanitycheck}

We create a lexicon of $\approx $ 73K words and their embeddings by using Word2Vec scheme described above. Since Turkish is an agglutinative language, it is not uncommon to come up with words with the same root but different suffixes in the lexicon. 

Before delving into further operations by using these embeddings, we want to perform a sanity check. In order to do this we randomly select some words  and find their similars by using the embeddings. The results can be seen in Table \ref{tab:sanitysimilarity}. 

\begin{table}[h!]
    \centering
    \begin{tabular}{|c|c|c|}
    \hline
        \textbf{Sample Word}   &   \textbf{Similar Word}&    \textbf{Similarity Ratio (SR)}\\    \hline  
        google  &   facebook&        0.7457 \\
                &   instagram&       0.7456      \\
                &   yandex&          0.6611      \\
                &   adwords&         0.6356      \\
                &   youtube&         0.6043      \\ \hline
        izmir   &   bayraklı&        0.8018      \\
                &   buca&            0.7978      \\
                &   alsancak&        0.7890      \\
                &   istanbul&        0.7813      \\
                &   balıkesir&       0.7730      \\ \hline
        ahmet   &   mustafa&         0.9792      \\
                &   murat&           0.9744      \\
                &   mehmet&          0.9666      \\
                &   ömer&            0.9627      \\
                &   hakan&           0.9593      \\ \hline                
        kazanç (winnings)  &   getiri (return)&             0.7059      \\
                            &   yarar (benefit)&            0.6985      \\        
                            &   yatırım (investment)&       0.6972      \\        
                            &   devamlılık (continuity)&    0.6704      \\                  
                            &   verimlilik (productivity)&  0.6666      \\                  
    \hline
    \end{tabular}
    \caption{Sample words and their close neighbours makes it possible to perform a basic sanity check for the obtained word embedding. }
    \label{tab:sanitysimilarity}
\end{table}

We can interpret the outcomes listed in Table \ref{tab:sanitysimilarity} as follows:

\begin{enumerate}
    \item \textbf{`google' : } The word `google' is one of the most commonly used words in the corpus since the company is providing consultancy mainly on Google Ads. Most similar five words, i.e. the words with the closest word embeddings, are found to be other popular internet advertising platforms (i.e. facebook, instagram, yandex etc.) for which the company also provides consultancy.
    \item \textbf{`izmir' : } `izmir' is the name of the city where the company is located in Turkey. First three matched words are popular district names in İzmir. The remaining two words (i.e. `istanbul' and `balıkesir') are other popular cities in Turkey. Since they are all indicating residential areas, the word embedding for `izmir' also seems consistent. 
    \item \textbf{`ahmet' : } `ahmet' is a very common male name in Turkey. Not surprisingly all five similar words are also common male names in Turkish language.  Hence embedding seems to capture not only the context but also gender of the words.
    \item \textbf{`kazanç (winnings)' : } For this specific example the English translations of the words are given in parentheses. As it can be seen from these samples, meaning of a relatively abstract word can also be captured with the trained word embeddings.
\end{enumerate}

\section{Case Studies}\label{sec:casestud}

In this section we introduce the case studies. First we try to perform text mining by directly using Gensim library without any further complex learning scheme. In addition to that, by using the labels (i.e. lead labels and agent ids) in the data we deploy two supervised learning algorithms which uses the word embeddings together with two slightly different bidirectional LSTM architectures.

\subsection{Text Mining}\label{sec:textmining}

In this case study we want to show some remarkable outcomes of directly using word embeddings for text mining:

\begin{itemize}
    \item We first find the most similar words of `adresgezgini' (i.e. name of the company). The corresponding function (i.e. most\_similar() function of Gensim library) returns a few of the competitor company names together with misspellings of the target word `adresgezgini'. When we further perform the similarity check by using the found competitor company names it  gives us some previously unknown competitor company names which only occurred a few times in the large corpus of notes. Hence, simple similarity check  scheme can mine out  very useful information from the corpus corpus of text notes.
    \item Similar to the previous application, we consider the word  `küfür' (swearing) which is a very negative word to have in a CRM note. The notes including this word are mostly about negative experiences occurred between the company and the customers. Hence it is  convenient to search for similar negative words in the notes as soon as they are saved to the database in order to take immediate actions to protect either the customers or the company employees. In Table \ref{tab:similartokufur} the list of similar words to `küfür' (and their English translations) are given. The second column indicates the number of detected notes which contains the corresponding word. The search with the word `küfür' returns only 82 cases where the search including its  first five alikes gives a total of 504 different cases (i.e. notes). Mining out these notes from within more than one million notes is an important outcome of directly using word embeddings.
\end{itemize}  
    
\begin{table}[h!]
    \centering
    \begin{tabular}{|c|c|c|}
    \hline
        \textbf{Word}&          \textbf{\#Cases  }&  \textbf{SR wrt.`küfür'}\\    \hline  
        küfür (swearing)&       82&     1.0\\
        hakaret (insult) &      64&     0.9200  \\
        tehdit (threatening)&   60&     0.9073  \\        
        taciz (abuse)&          18&     0.9054  \\        
        isyan (revolt)&         30&     0.9043  \\                  
        itiraz (objection)&     250&    0.9031 \\                  
    \hline
    \end{tabular}
    \caption{The list shows the number of detected notes by performing SQL searches using the word `küfür' and the words which has closest meaning to it.}
    \label{tab:similartokufur}
\end{table}

\subsection{Lead Labelling}\label{sec:leadlabelling}

In this section we want to present the solution to a problem called ``lead labelling" which is important for time management since it is not desired by the companies to  spend time on low priority  leads rather than high priority leads. In the company, lead labelling is performed by the managers. They have to carefully read the notes taken by the agents and decide which priority  (i.e. label) is to be assigned to the corresponding lead. This is a very slow process and it is not possible to read every single note and prioritize each lead in a timely manner. But once implemented, an AI algorithm can automatically make this segmentation by analysing every note as soon as they are stored in the CRM database.

\begin{figure}[h!]
  \centering
  \includegraphics[width=0.50\textwidth]{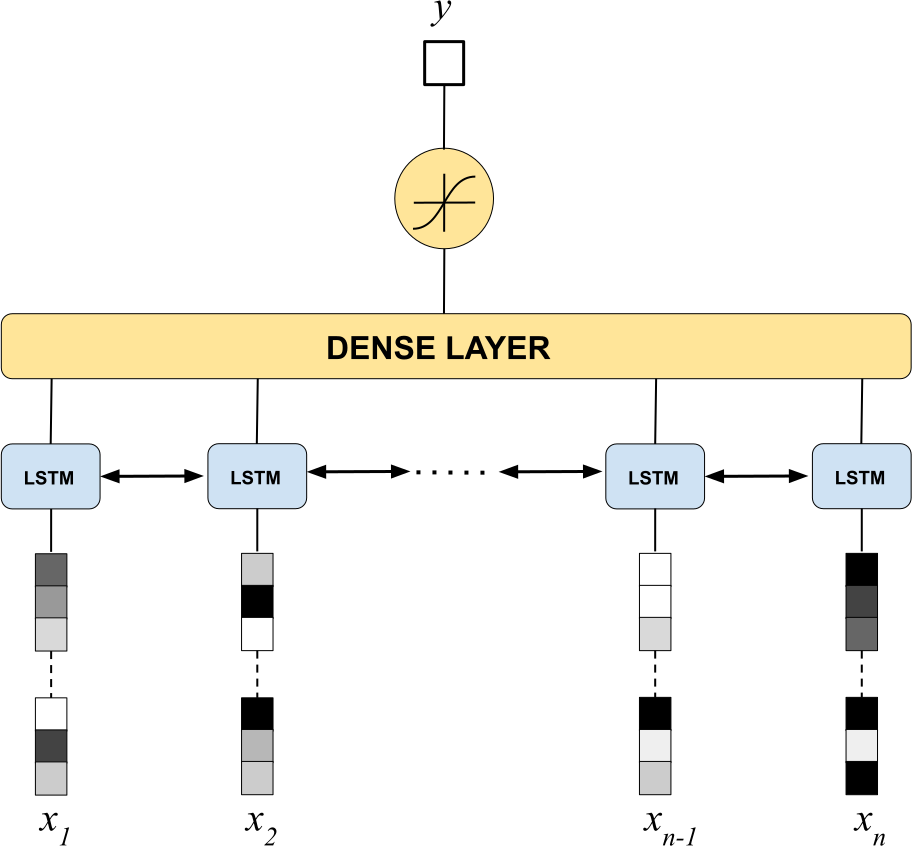}
  \caption{RNN architecture constructed by using $n$ cascaded bidirectional LSTM units. The output of dense layer is connected to a sigmoid output which is capable of performing binary segmentation.}
  \label{fig:nwbinary}
\end{figure}

The proposed algorithm is expected to mimic the behaviour of the managers which requires capturing the meanings of the notes. It is a convenient approach to use bidirectional recurrent neural networks for such NLP problems. The input of proposed network is going to be the sentences and its output is going to be either 1 (high priority label) or 0 (low priority label) which reduces problem down to a binary classification problem. The output labels of the training set are obtained by using  a priori segmentation information which is manually created by the managers. We can depict the architecture as Figure \ref{fig:nwbinary}. 

In this figure the input is a series of words. We selected the number of words for a single training input (i.e. $n$) as 100. The words ($x_1 \dots x_n$) are vectors obtained from Word2Vec representation scheme. We repeated the experiments for different embedding sizes and the corresponding results will be given in Section \ref{sec:results}. 

Input layer is followed by a bidirectional LSTM layer of $n$ units which is connected to a dense layer. Since the problem is constructed as to perform a binary classification for the leads in the CRM, the output layer is a single sigmoid node.

\subsection{Agent Identification}\label{sec:agentidentification}

Author identification is an active research subject where modern ML methods are highly appreciated. In \cite{Rehman:2018} this problem and most recent solution methods are explicitly addressed. In this work we slightly modify the previous architecture which we use for lead labelling problem and apply it to author identification problem since using LSTM architecture  is a  convenient method also for this problem \cite{Gupta:2019}.

We wanted to see whether the data which is already available to perform a binary classification for leads can be used for a much complex classification problem like author identification or not. As we further analysed the data we found out that, as a matter of course, there is much more data entered by more senior agents compared to junior ones. By calculating word counts for each agent and selecting the agents who already entered more than 300K words, a training set for identifying  14 different agents is created. The scheme can be extended such that it includes every single agent. But especially for display purposes we kept the number of classes limited to 14.

\begin{figure}[h!]
  \centering
  \includegraphics[width=0.5\textwidth]{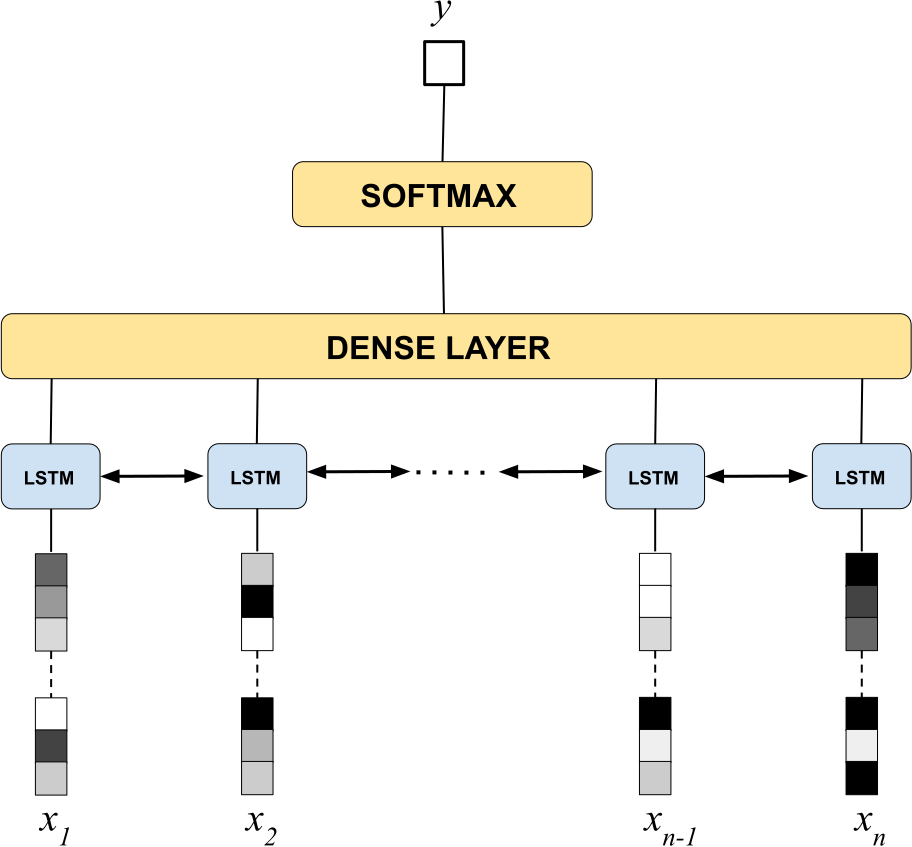}
  \caption{This architecture is the same as the architecture depicted in Figure \ref{fig:nwbinary} except  the output layer. Since this network is to be used for multiclass classification a softmax layer is used instead of a single sigmoid node.}
  \label{fig:nwmulticlass}
\end{figure}

The architecture in Figure \ref{fig:nwmulticlass} for agent identification is very similar to the binary classification architecture in Figure \ref{fig:nwbinary} except for the output layer. The sigmoid layer in the previous problem is replaced with a softmax layer since the problem now becomes a multiclass classification problem.  

In Section \ref{sec:results} we show the identification results for the 14 selected agents and the results show that if there is more data its more likely to identify an agent better from given text segments. Since we obtain convincing results from agent identification process it makes us to further ask can the same architecture be used to predict the next word of an agent while inputing a text, since LSTMs are very effective in learning complex language related tasks \cite{Karpathy:2015}. 

We slightly modified the structure in Figure \ref{fig:nwmulticlass} and We kept the input as small as three words ( i.e. $n=3$ in Figure \ref{fig:nwmulticlass}). The output softmax layer is adjusted in such a way to predict the next word from the dictionary. The training data is prepared again by using the agent notes. Assuming that a note is a sequence of words like $(w_1,\dots,w_n)$ we can select three consecutive words $(w_i,w_{i+1},w_{i+2})$ as training input where the next word, i.e. $w_{i+3}$, becomes the output. We also show the results of this language model  in Section \ref{sec:results}.

\section{Implementation Details}\label{sec:implement}

We dedicate this section for the details of implementation. For both prototyping and deploying we used Python \cite{Rossum:1995}. Google Colab platform is a free platform where one can natively use Jupyter Notebooks and Python. If one is planning to use the paid Google Cloud Platform (GCP) service for a deep learning task, it is convenient to perform prototyping, and even trying a few epochs of training on free GCP service then migrate the code to GCP. GCP also gives an opportunity to use a single graphical processing unit (GPU)  for free.

For the first case study in Section \ref{sec:textmining}, Gensim libary is used to generate word vectors. This library can perform iterative Word2Vec training without much effort. After preprocessing the raw text data according to the needs, it can directly be used  for Word2Vec training. It creates a dictionary of different words used in the corpus. It is possible to filter out the words which occur less than a specific number (i.e. 5 in our case) of times. It is also possible to specify the number of iterations. We found out that 5 to 10 iterations are sufficient for our case.

Gensim library also gives you the opportuity to set the size of word embeddings. According to \cite{Jurafsky:2019}, the size of the word embeddings is crucial to prevent overfitting since dense embedding vectors tend to overfit less than sparse vectors. In order to make a thorough analysis of word embedding size , we used the embedding sizes 50, 100, 200 and 300 respectively. The results will be discussed in Section \ref{sec:results}.

In order to implement the RNN architectures, Tensorflow \cite{Tensorflow:2015} is preferred. One of the most crucial and time consuming process of such ML problems is to tailor the data for training. But thanks to such libraries like Scipy \cite{SciPy:2008}, Numpy \cite{Numpy:2008} and Pandas \cite{Pandas} it is relatively easy to perform necessary preprocessing operations. 

Two slightly different versions of the same RNN architecture are used in the case studies in Sections \ref{sec:leadlabelling} and \ref{sec:agentidentification}. Both architectures require an embedding layer in the input since in the training data set we only keep the index number of each word. The size of the embedding matrix for different vector sizes is $(73.310,n)$ where $n$ is the embedding size and $73.310$ is the number of words in the dictionary. Even we select the smallest size (e.g. 50) for a word vector, it is not memory efficient to use the embedding vectors in the training data set directly. Hence the embedding layer is important and it is easily handled in Tensorflow once the embedding matrix is correctly constructed. 

The embedding matrix also includes a tag  for both unknown words  ($<UNK>$) and padding tag ($<PAD>$). Padding is used to complete a short training input data size to the default size. In this study we prefer the input size to be 100, which means that the sentences we used for training can have at most 100 words. The sentences with less than 100 words are padded with $<PAD>$ tags and sentences with more than 100 words are trimmed from the end to obtain the correct size. 

The core layer of both architectures is the LSTM layer. We created a bidirectional LSTM with 64 hidden nodes where output of the layer is fully connected to  a dense layer of 256 nodes with ReLU activation function \cite{Vinod:2010}.

Two classification architectures only differ in the output layer. For the binary classification problem in Section \ref{sec:leadlabelling} it is only a single sigmoid node but for the multiclass classification problem it is a softmax layer which is, for this specific problem, capable of classifying the data into 14 different classes. One other major implementation difference between two problems is that the specified loss function for the former problem is chosen as binary cross entropy (BCE) and for the latter one as categorical cross entropy (CCE). They both measure the overall difference between the target and predicted labels of a classification problem. BCE can be formulized as Equation \ref{eq:bce} for  target ($y$) and predicted ($\hat{y}$) labels where N is the number of training samples. CCE can similarly be formulized as Equation \ref{eq:cce} where, in addition to N, M is the number of different labels.

\begin{align} 
L_{BCE}(y,\hat{y}) &= - \frac{1}{N} \sum_{i=0}^N(y*log(\hat{y}_i) + (1-y)*log(1-\hat{y}_i)) \label{eq:bce}\\
L_{CCE}(y,\hat{y}) &= -\sum_{j=0}^M\sum_{i=0}^N(y_{ij}*log(\hat{y}_{ij})) \label{eq:cce}
\end{align}

More to emphasize, in both training phases ADAM optimizer \cite{Kingma:2015} is used. For the lead labelling and agent identification problems we use 1024 and 512 batch sizes respectively. We also used latter scheme to perform word prediction for a specific agent, where each four consecutive (first three for input and last one for output) words in agent notes are used as training data. The output layer is still a softmax layer but in this case it is as big as the dictionary size of the corresponding agent which is about 20K words for the selected agent.

\section{Results and Discussion}\label{sec:results}

In this section we show the experimental results of the case studies. The first case study in Section \ref{sec:textmining} was a straightforward application of the \texttt{most\_similar()} function of Gensim library to the text corpus and finding similar words for any selected word. The outcomes of these experiments were explicitly given in the corresponding section.

The latter case studies were about semantic classification of given text segments according to training data set. We used the text corpus and created a training data set where the input is any note stored in CRM and the output is either the binary label (1 or 0) for lead labelling problem or one of the 14 agent ids for agent identification problem.

\begin{figure}[h!]
  \centering
  \includegraphics[width=1.0\textwidth]{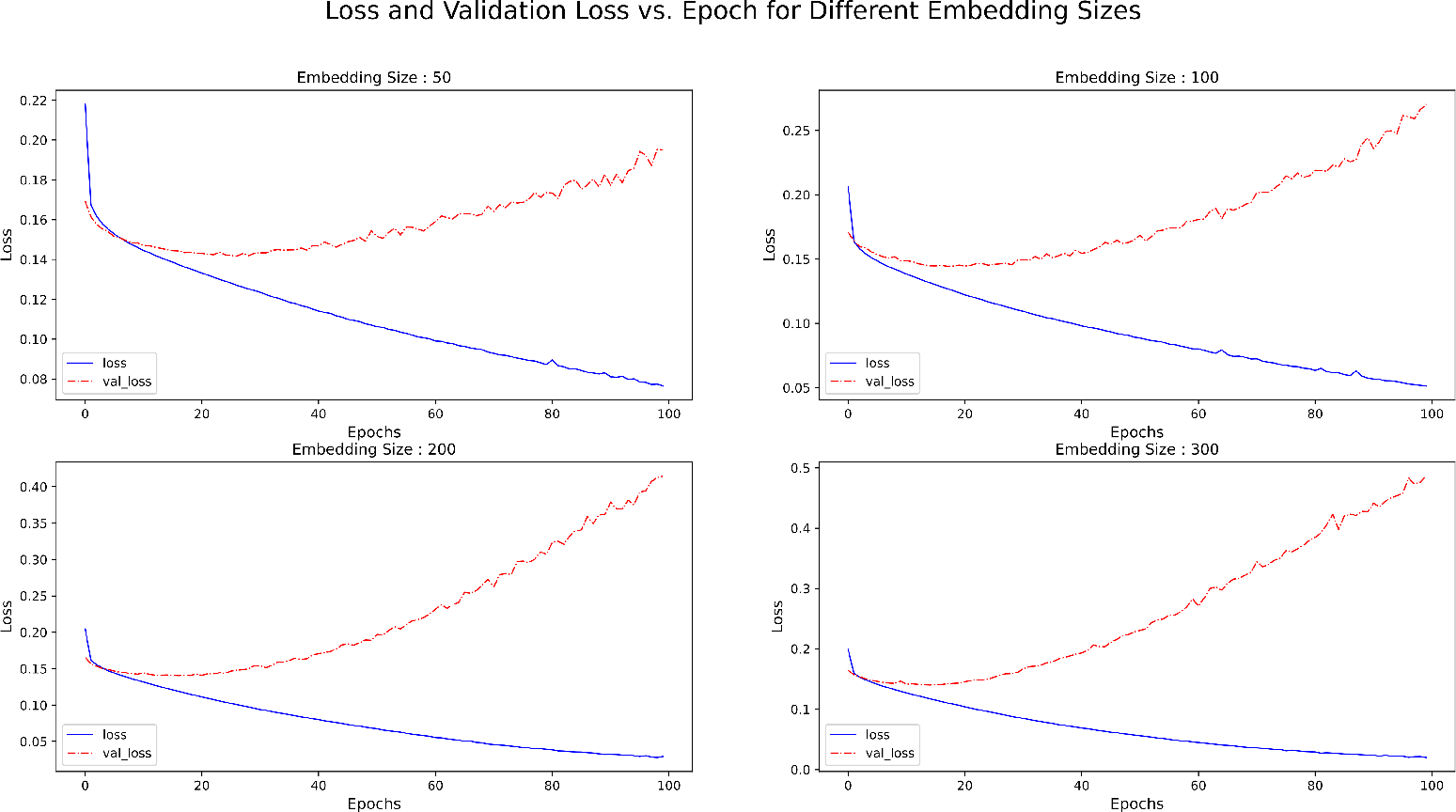}
  \caption{The training and validation loss plots for different embedding sizes 50, 100, 200 and 300 respectively. Each experiment is continued for 100 epochs.}
  \label{fig:lossvalloss}
\end{figure}

For the lead labelling case study we trained the architecture depicted in Figure \ref{fig:nwbinary} for 4 different embedding vector sizes. For each experiment we continued training for 100  epochs. We used \%90-\%10 split scheme for training and validation data sets. The corresponding training and validation loss plots can be seen in Figure \ref{fig:lossvalloss}.

\begin{figure}[h!]
  \centering
  \includegraphics[width=0.6\textwidth]{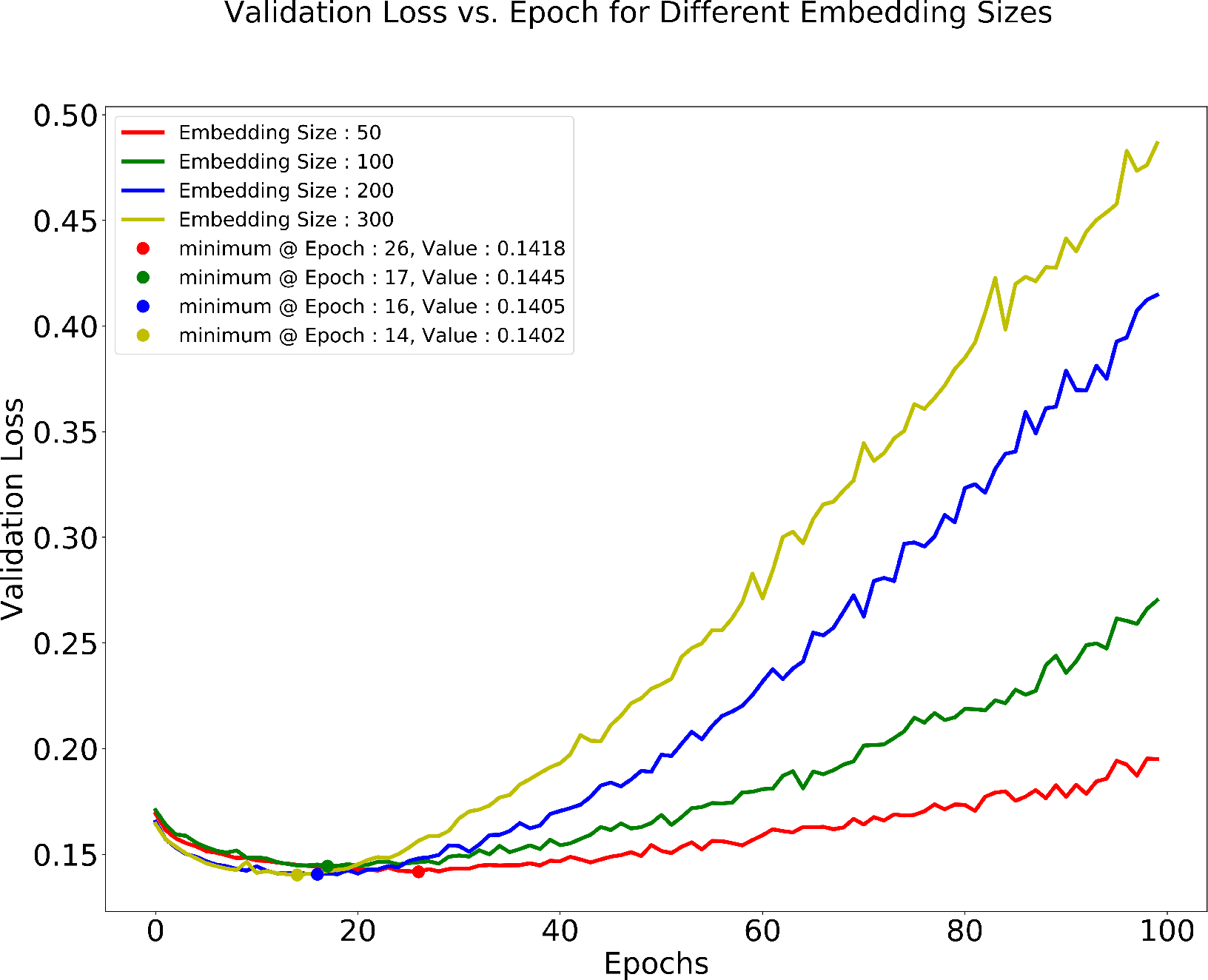}
  \caption{The validation losses for different embedding sizes in Figure \ref{fig:lossvalloss} are plotted together. The dots indicate the epoch where the validation losses reach their global minimum values.}
  \label{fig:valloss}
\end{figure}

The common problem with these graphs is that the validation loss first decreases and then starts to increase at some point. This increase is a side effect of the overfitting problem. If we plot only the validation losses together in the same plot we obtain Figure \ref{fig:valloss}.

 If we interpret Figure \ref{fig:valloss} it can be seen that for each embedding size validation loss function plateau down and soon after reaching the global minumum it starts to increase.  The minimum values are very close, but the epoch where the system starts to overfit differs. Recalling the fact that sparse embedding vector representations tend to overfit, we may expect to have premature overfitting for bigger embedding sizes. This claim can also be proven by looking at the slopes of the curves in Figure \ref{fig:valloss}. Moreover the validation loss starts to increase after 26th epoch for embedding vector size 50 and after 14th epoch for embedding vector size 300. Hence for denser embedding vectors it takes longer time for the system before it starts to overfit.
 
 To decide which embedding size is more preferable, we have to look at another metric which reflects the accuracy. F1 Score is much appreciated for binary classification problems since it captures the precision and recall inherently. Precision, Recall and F1 Score can be given as Equations \ref{eq:precision}, \ref{eq:recall} and \ref{eq:f1} where TP, FP and FN are true positive, false positive and false negative values respectively.
 
\begin{align}
    \text{Precision} &= \frac{TP}{TP+FP}\label{eq:precision}\\
    \text{Recall} &= \frac{TP}{TP+FN}\label{eq:recall}\\
    \text{F1} &= 2*\frac{\text{Precision}*\text{Recall}}{\text{Precision}+\text{Recall}}\label{eq:f1}
\end{align}
 
\begin{figure}[h!]
  \centering
  \includegraphics[width=0.6\textwidth]{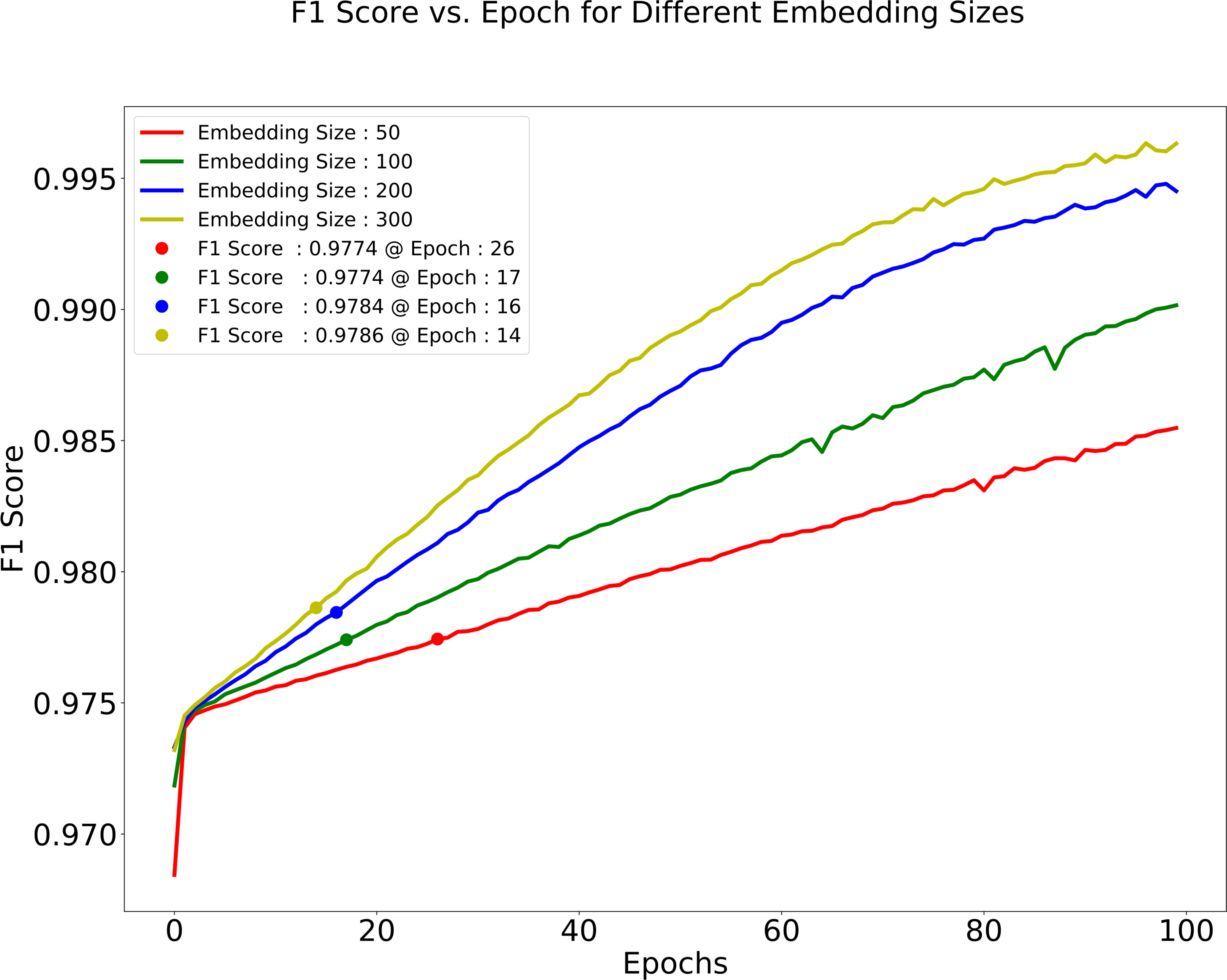}
  \caption{F1 Scores change for the training experiments in Figure \ref{fig:lossvalloss} shown in the same plot. The dots indicate the epochs where the system starts to overfit.}
  \label{fig:f1}
\end{figure}

 If  F1 Score for 100 epochs is plotted for different embedding vector sizes we obtain Figure \ref{fig:f1}. The dots correspond to the epoch time where the system starts to overfit. These points are important to decide where to stop training to prevent overfitting. This is called ``early stoppping" and this approach is commonly used in gradient descent learning \cite{Yao:2007}. Since the main objective of deep learning is to minimize the loss function using gradient descent, early stopping is  an accepted strategy also in deep learning literature. The colored dots in Figure \ref{fig:f1} are determined by using early stopping condition which points out the epoch where validation loss reaches its minimum and starts to increase. Four different epochs 26, 17,16 and 14 are shown for embedding sizes 50, 100, 200 and 300 respectively. 
 
 The achieved F1 Scores for corresponding epochs have very close values. Since it is better to have as much training epoch as possible, embedding size can be chosen as 50 with which the system can be trained longer before it overfits. Assuming that we stop at the indicated early stopping conditions the required time for the system training can be seen in Table \ref{tab:trainingtime}. The hardware configuration we used for the complete training epochs on GCP is a notebook instance with 1 Tesla K80 GPU and 4 vCPUs with 15 GB memory. 
 
\begin{table}[h!]
    \centering
    \begin{tabular}{|c|c|}
        \hline
        \textbf{Embedding Size}&   \textbf{Training Time}\\
        \hline
        50 (@ epoch 26)&  2247 secs.\\
        100 (@epoch 17)& 1773 secs.\\
        200 (@epoch 16)&  1990 secs.\\
        300 (@epoch 14)&   2021 secs.\\    
        \hline  
    \end{tabular}
    \caption{Time required to train the system architecture in Figure  \ref{fig:nwbinary} for different word embedding dimensions up to specified epoch with respect to early stopping condition. }
    \label{tab:trainingtime}
\end{table} 

\begin{figure}[h!]
  \centering
  \includegraphics[width=0.6\textwidth]{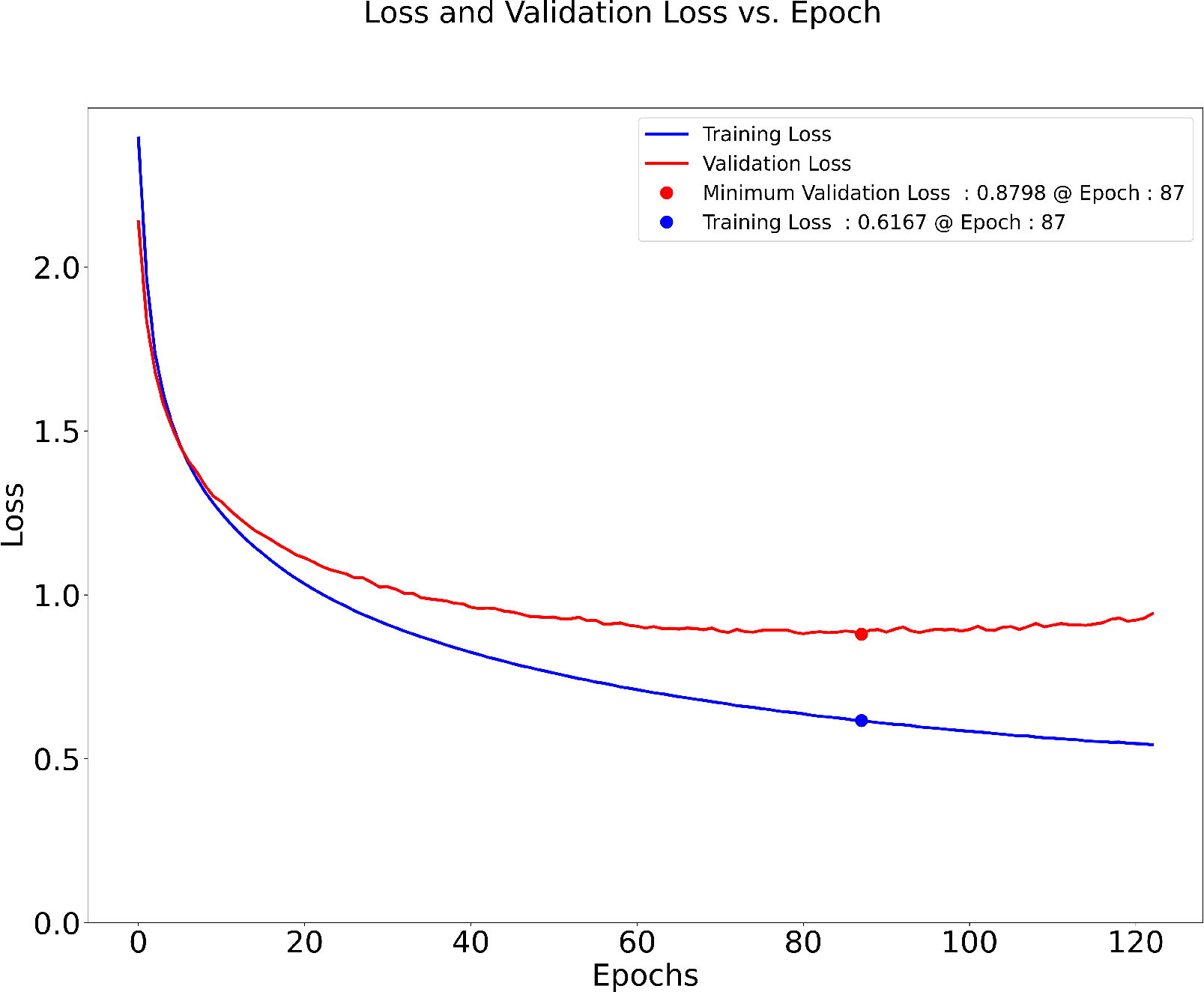}
  \caption{Training and validation loss vs. epoch plot for agent identification problem. The dots indicate the epoch where the validation loss reaches its global minimum and the system starts to overfit.}
  \label{fig:aslossvalloss}
\end{figure}

By using the outcomes of lead labelling problem, we decided to use word embedding size of 50 for the agent identification problem as well. 14 senior most agents' ids  and their notes (total of $\approx 236K$ notes) are selected for training. After training the system for 125 epochs we obtained the joint training and validation loss plot in Figure \ref{fig:aslossvalloss}.

If we use the early stopping condition we obtain 87th epoch where the validation loss starts to increase and the system starts to overfit.

\begin{figure}[h!]
  \centering
  \includegraphics[width=0.6\textwidth]{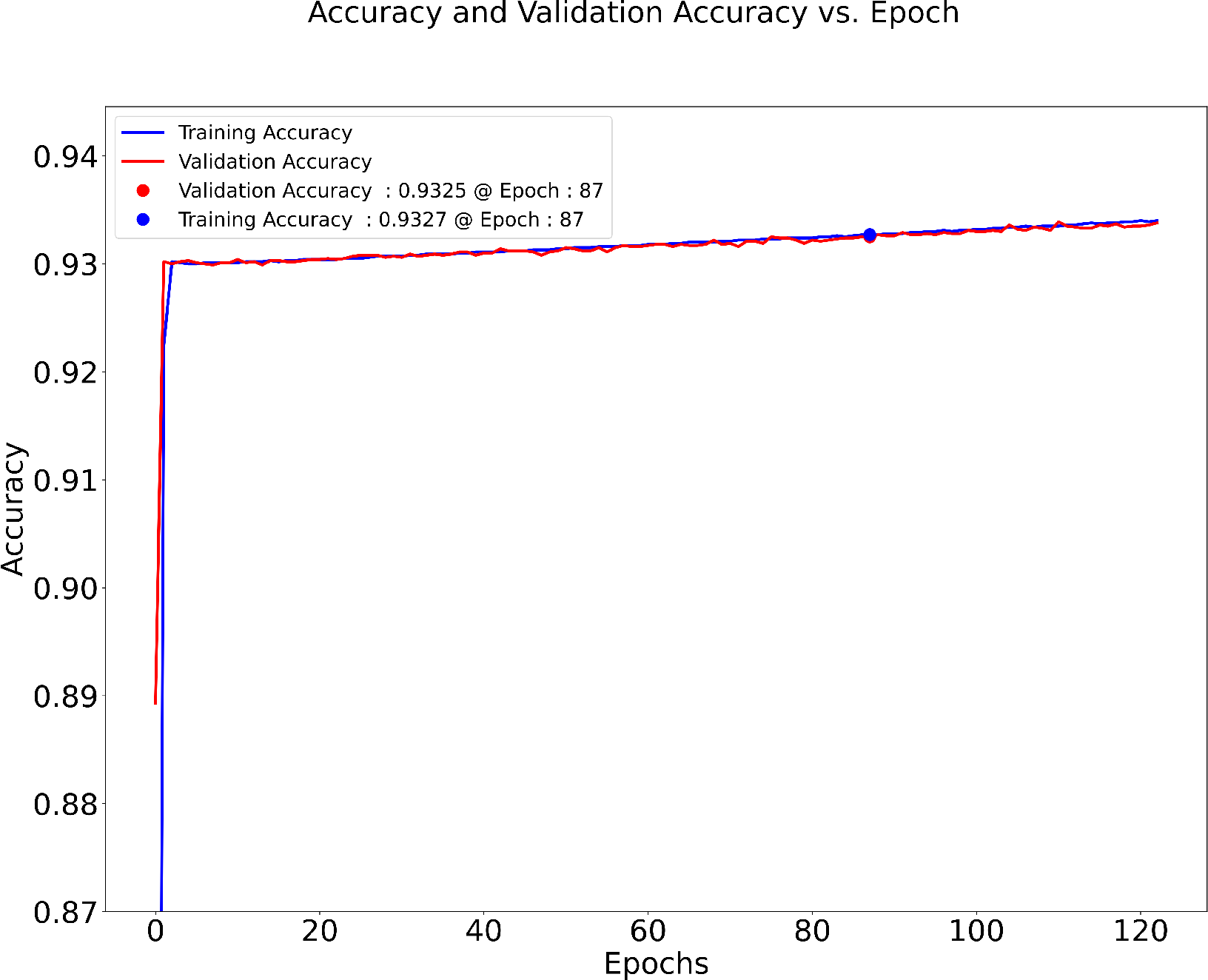}
  \caption{Training and validation accuracy vs. epoch for agent indetification problem. The dots point the early stopping condition occured at 87th epoch.}
  \label{fig:asaccvalacc}
\end{figure}

\begin{figure}[h!]
  \centering
  \includegraphics[width=0.7\textwidth]{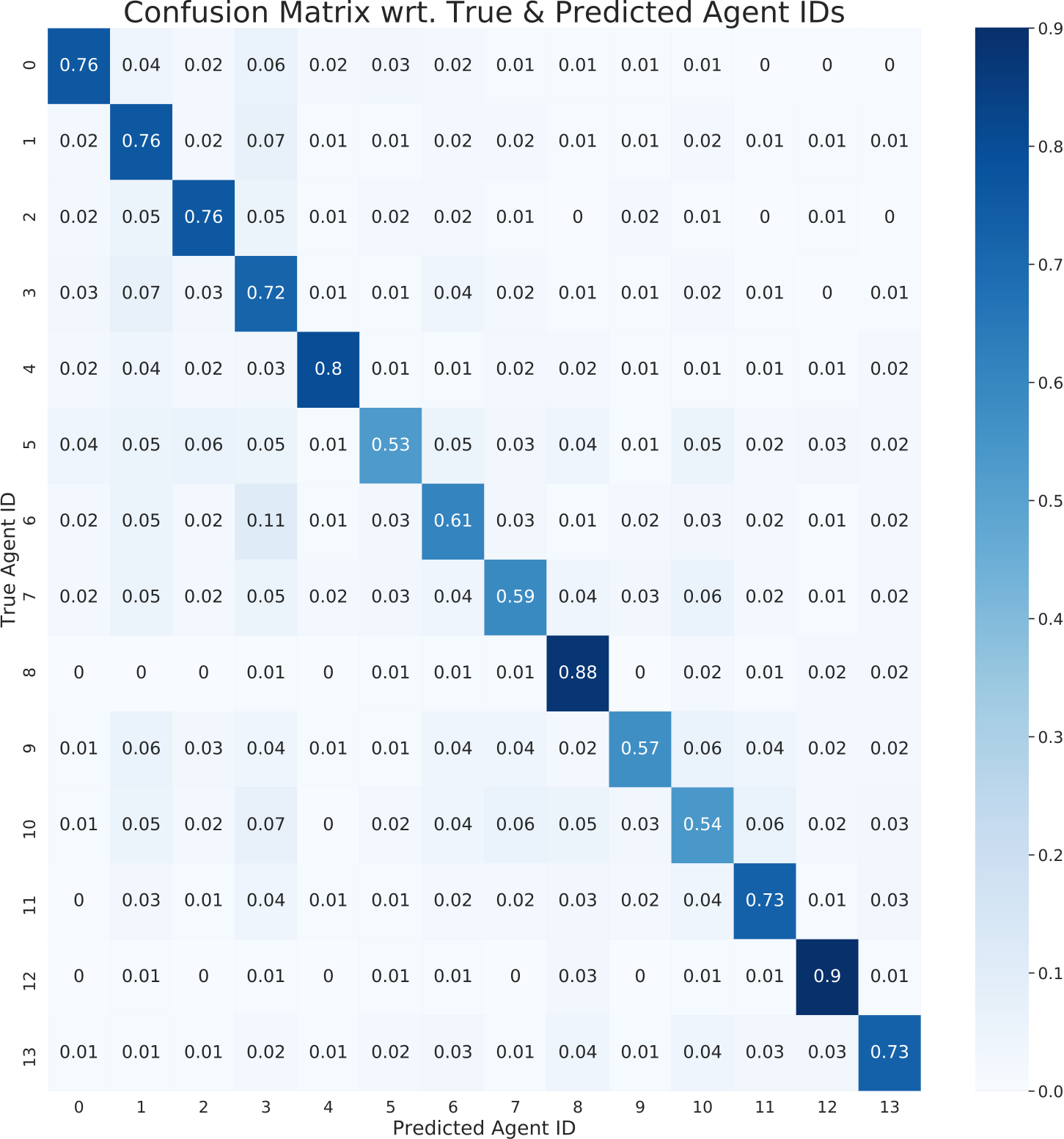}
  \caption{The confusion matrix which is created by finding the classification error of the agent identification problem introduced in Section \ref{sec:agentidentification}.}
  \label{fig:confmatrix}
\end{figure}

By plotting the overall accuracy of the system both for training and validation data sets we obtain Figure \ref{fig:asaccvalacc}. At 87th epoch the system achieves about \%93 accuracy both for training and validation data sets. But it is important to examine misclassifications of the system. Confusion matrices help to see the classification errors visually. The confusion matrix for this learning scheme can be seen in Figure \ref{fig:confmatrix}. This matrix is generated by using the validation data set. 

For an ideal classification (i.e. \%100 accuracy) the diagonal values of the confusion matrix become 1.0. But in practice this is not possible. If we examine the confusion matrix values in Figure \ref{fig:confmatrix} we can see that some values are very close to 1.0 and some values are below 0.70. When we further investigate the agent ids corresponding to lower diagonal scores we see that these are the agents with less data samples. Similarly the best diagonal scores are mostly related with the agents who have more data samples in the training data set. 

We expanded this scheme and further train it as to predict the next word for given set of n words where n is 1,2 or 3. This scheme can be trained for each agent separately such that the network also learns writing style of a specific agent. In our experiments we trained the network with a specific agent's notes in the database. The performance of the training is convincing such that after training for about 300 epochs,  it predicts the next word within the first three most probable guesses with \%72 accuracy.

\subsection{Conclusions}\label{sec:concs}

In this study we proposed three case studies about making a good use of the text corpus stored in a CRM database by using state of the art methods used in NLP literature. The data used in the experiments is the customer related notes taken between the years 2009 and 2020 by the agents of an internet ads consultancy agency located in Turkey. There are about 1 million notes, each of which is related with a unique customer id and a unique agent id, in the company's database . Furthermore, each customer is assigned with either a high priority or low priority label. 

In the first case study we first trained word embeddings by using Word2Vec method. The training yielded a lexicon of approximately 73K words. The word embeddings intrinsically enables us to find semantically similar words to a given word according to the training data. We showed two useful applications of this scheme in Section \ref{sec:textmining}. Such an implementation can be automatised to train itself occasionally and detect entries having negative words just by giving a few samples and populate the search list with \texttt{most\_similar()} function of Gensim library. It'd be a good approach to further implement a search bar in the CRM which captures the entries including the keyword and also the entries which includes the most related keywords found by the proposed method. Such an application can make managers to detect possible problems happening during CRM processes effectively.

The remaining two case studies were about classification which required supervised learning. We applied the word embedding intuition also to these problems by using it together with LSTM-RNN architecture.  The training data is of grave importance for supervised learning in a deep learning. It should be consistent and be appropriate for the problem solution. After creating the appropriate data for both problems, the training experiments were performed. 

For lead labelling case we tried 4 different word embedding sizes in order to determine the best size. After the experiments we figured out that 50 is an ideal size for our training data. For different data sets the result may vary, hence the optimal word embedding size should be found by testing various sizes as we presented in Section \ref{sec:leadlabelling}. We trained the network for 100 epochs and verified the claim that sparser word vectors are prone to overfit but with smaller vector size the system can be trained longer. Since the achieved accuracy values are very close, we adopt the smallest vector size, 50, as the embedding size for both problems. 

Lead segmentation is an important managemental strategy in order to ensure time management. Especially companies which have to deal with numerous cases simultaneously, it is important to prioritize CRM tasks and lead segmentation is one of them. In terms of time management, it is now inevitable for the companies to leave some of the human dependent operations to computers. Fortunately recent advances in both hardware technology and AI literature now makes these kinds of transformations possible. In the corresponding case study we presented how we applied recent NLP methods to replace the company's binary lead segmentation strategy. Same strategy can be extended to perform this segmentation for more number of classes if there exist sufficient training data regarding different classes. 

In order to solve the agent identification problem in Section \ref{sec:agentidentification}, we inherited the same architecture we used for lead labelling. The only difference was the output layer since the problem became  a multiclass classification problem. The deep learning solution of this problem yielded \%93 accuracy in correctly identifying 14 different agents ids from the notes selected from the CRM database. When we further analysed the results we figured out that most of the classification errors are due to the lack of data, i.e. the system identifies an agent better as the system populates more data for that specific agent. With the proposed scheme any CRM can inherently gain the feature of identifying its users. We also expanded the same scheme in such a way that it can predict a specific agent's next word for given few input words. This prediction system can be used during note taking processes for agents once it is trained with each agents entries separately. By using the same word level training logic, the network can figure out an agents next word for a priori given three consecutive words. Such an application would make a CRM smart enough to speed up and ease agents text entry processes. Furthermore, such an application can make it easier for the agents to enter a text in a CRM especially via a mobile device interface.

Most of the professional CRM software companies in the market have been using NLP in their services and products for a while. But this study shows that, any company can upgrade its own CRM in many ways by using conveniently implemented NLP algorithms for their own needs without paying for a costly third party CRM software.

\bibliographystyle{spmpsci}      
\bibliography{mybibfile}   

%
%

\end{document}